\documentclass{article}
\usepackage{iclr2027_conference,times}
\usepackage[utf8]{inputenc}
\usepackage[T1]{fontenc}
\usepackage{hyperref}
\usepackage{url}
\usepackage{booktabs}
\usepackage{amsmath,amsfonts,amscd,amsbsy}
\usepackage{nicefrac}
\usepackage{microtype}
\usepackage{caption}
\usepackage{graphicx}
\usepackage{subcaption}
\usepackage{multirow}
\usepackage{latexsym,bm}
\usepackage{mathtools}
\usepackage{euscript}
\usepackage{makecell}
\usepackage{textcomp}
\usepackage{enumitem}
\usepackage{inconsolata}
\usepackage{xspace}
\usepackage{svg}

\usepackage{wrapfig}
\usepackage{graphicx}

\setenumerate[1]{itemsep=0pt,partopsep=0pt,parsep=\parskip,topsep=5pt}
\setitemize[1]{itemsep=0pt,partopsep=0pt,parsep=\parskip,topsep=5pt}
\setdescription{itemsep=0pt,partopsep=0pt,parsep=\parskip,topsep=5pt}

\newcommand{\minisection}[1]{\vspace{0pt}\noindent\textbf{#1.}}
\newcommand{\modelname}{MCPDisCo\xspace}

\title{\modelname: Seesaw-Mitigated Tuning for MCP-Native Agents}

\author{Kounianhua Du$^1$, Jianxing Liu$^1$, Shuai Shao$^1$, 
 Ke Li$^2$, Xing Sun$^2$, Weinan Zhang$^1$\\
 $^1$Shanghai Jiao Tong University, $^2$ Tencent.\\
Shanghai, China\\
\{kounianhuadu,flweb3ranni,shaoshuai.ederson,wnzhang\}@sjtu.edu.cn \\
\{tristanli,winfredsun\}@tencent.com}

\iclrfinalcopy

\begin{document}
\maketitle
\begin{abstract}
The Model Context Protocol (MCP) provides a shared interaction
substrate for the agentic web, allowing heterogeneous services
to expose capabilities through a standardized interface.
Learning to use this ecosystem entails a hierarchy of
capabilities: tool-format skills, server-level transfer,
and cross-server composition.
We present \modelname, an ecosystem-scale reinforcement
learning framework for MCP-native agents.
Using a Thought-Markov Decision Process (Thought-MDP),
the framework explicitly represents internal reasoning
alongside routing, execution, and termination, and trains
on real-server, long-horizon interactions spanning
70 MCP servers and 527 tools across six domains.
We study the reasoning--tool seesaw, a joint-learning
failure mode in which improvement in one objective
can accompany deterioration in the other.
To mitigate potential interference, \modelname combines
token-masked reward attribution with bidirectional gradient
surgery: the former separates direct supervision for
reasoning and action spans, while the latter removes
conflicting components from both objective gradients
in shared parameters.
\modelname achieves higher success with \emph{fewer} chain-of-thought tokens. On the out-of-distribution MCP-Atlas and GAIA benchmark, \modelname maintains monotone gains, confirming that the learned behavior generalizes beyond the training server distribution and broader agentic tasks. Model weight and implementation are available. \footnote{https://anonymous.4open.science/r/MCPDisCo/README.md}.

\end{abstract}

\section{Introduction}

Large language models (LLMs) are increasingly deployed as agents
that retrieve information, invoke tools, and coordinate external
services to complete user tasks
\citep{schick2023toolformer,yao2023react,qin2024toolllm}.
As these tasks extend across applications, effective agency
requires more than generating a valid function call.
An agent must identify relevant capabilities, determine how
to use them, and adapt its plan to intermediate observations.
Supporting these behaviors across heterogeneous services
requires both a common interaction interface and policies
that generalize beyond familiar tools and workflows.

The Model Context Protocol (MCP)~\citep{mcp} provides a
standardized client--server interface through which external
services expose their capabilities to agents.
We position MCP as a shared interaction \emph{substrate for
the agentic web}: a common foundation on which independently
developed capabilities can be accessed and combined into
task-oriented workflows.
By separating protocol-level integration from the underlying
service implementation, MCP reduces the need for bespoke
connectors. Combined with server discovery and agent-side
orchestration, it enables agents to locate relevant
capabilities and compose their use across services.
Its significance therefore extends beyond standardizing
individual tool calls to supporting an extensible environment
for agentic interaction.

This perspective motivates a hierarchy of generalizable
capabilities: \emph{tool-format skills},
\emph{server-level transfer}, and
\emph{cross-server composition}.
MCP interaction data covers the
\emph{query--server--tool} chain, embedding tool invocation
within server discovery, capability selection, and execution
feedback. Conventional tool calling is thus a component of
this richer interaction structure.
This motivates an \emph{asymmetric generalization hypothesis}:
learning from MCP interactions can support conventional
tool-use tasks, whereas training on individual tool calls
alone need not teach the server-level decisions required
for MCP tasks.
Moreover, standardized server interfaces allow capabilities
to be wrapped and recombined into diverse workflows.
Real-server, long-horizon data can exploit this flexibility
to expose agents to complex task compositions and sustained,
observation-conditioned decision making.
The hierarchy offers a structural basis for transfer,
rather than a guarantee that MCP training will outperform
other training regimes on every downstream task.

Recent work has begun to develop the evaluation and data
infrastructure for MCP-native learning.
LiveMCPBench~\citep{mo2025livemcpbench} and
MCP-Atlas~\citep{bandi2026mcp} evaluate agents using real
MCP servers, while MCP-Flow~\citep{wang2026mcpflow}
constructs large-scale MCP training data and studies
supervised tool learning.
These efforts establish important foundations, but learning
to coordinate reasoning, discovery, and execution over
extended interactions also raises an optimization question:
how should an agent jointly improve its reasoning and
tool-use behavior?
We investigate this question through reinforcement
learning (RL) on real-server, long-horizon MCP tasks,
complementing existing benchmark and data-centric efforts.

Reasoning and tool use are complementary at inference time,
but their training objectives need not improve together.
We refer to a trade-off in which progress on one objective
is accompanied by deterioration on the other as the
\emph{reasoning--tool seesaw}.
Related work also studies reasoning--tool interference
in agentic RL~\citep{li2026reasoning}.
Two forms of coupling are relevant in our setting.
First, assigning a combined reward to every generated token
gives reasoning and action spans the same direct supervision,
despite their distinct roles.
Second, even separately attributed rewards update shared
parameters and can induce opposing gradient directions.
Gradient interference is a potential mechanism underlying
the seesaw, rather than equivalent to the behavioral
trade-off itself.
Addressing these two forms of coupling requires interventions
at both the token-attribution and parameter-update levels.

We introduce \modelname, an ecosystem-scale RL framework
for MCP-native agents.
We instantiate a Thought-Markov Decision Process
(Thought-MDP) that explicitly represents internal reasoning
alongside routing, execution, and termination.
This formulation supports reasoning-enabled interaction,
in which the agent deliberates over available capabilities
and uses returned observations to guide subsequent actions.
Training uses real-server, long-horizon tasks spanning
70 heterogeneous MCP servers and 527 tools across six domains.
To mitigate reasoning--tool interference, we combine
\emph{token-masked reward attribution}, which separately
supervises reasoning and action spans, with
\emph{bidirectional gradient surgery}, which removes
conflicting components from both objective gradients.
Neither objective is a privileged anchor.
The resulting framework targets task-relevant, coherent
reasoning alongside correct tool use, without introducing
an explicit reasoning-length penalty.

Experiments show that \modelname increases LiveMCPBench
task success from 15.79\% to 26.32\% while reducing average
chain-of-thought tokens by 5.6\% relative to the untuned
backbone.
Component ablations and gradient diagnostics examine the
effects of masking and surgery on task performance and
optimization interference.
On MCP-Atlas, \modelname outperforms the untuned backbone
and the evaluated tool-tuned models in both accuracy and
mean coverage, supporting transfer to unseen servers.
In GAIA~\citep{mialon2024gaia}, it outperforms ToolACE-8B
and Hammer-7B at Levels~1 and~2, while improvement over
the untuned backbone is restricted to Level~1.
These cross-benchmark results are consistent with the
proposed transfer asymmetry: MCP-native learning supports
broader agentic tasks, whereas conventional tool-call
tuning alone remains insufficient for strong MCP performance.
They do not, however, isolate the effects of data hierarchy
from those of the training algorithm.

Our contributions are fourfold:
\begin{itemize}[leftmargin=10pt]

\item \textbf{Positioning MCP as the substrate of the agentic web and showing the generalization hierarchy.}
We connect MCP's standardized interface to a generalization
hierarchy spanning tool-format skills, server-level transfer,
and cross-server composition.
The query--server--tool structure motivates asymmetric
transfer between MCP-native and conventional tool-use
learning, while flexible capability composition supports
complex agentic workflows.
We examine this perspective through evaluations on
MCP-native and broader agentic benchmarks
(Section~\ref{sec:ood}).

\item \textbf{Reasoning-enabled Thought-MDP and ecosystem-scale MCP-native RL.}
We instantiate a Thought-MDP framework that explicitly
supports reasoning alongside routing, execution, and
termination, and develop \modelname for RL tuning on
real-server, long-horizon tasks spanning 70 servers,
527 tools, and six domains.
The resulting policy improves task success over the
untuned backbone and evaluated tool-tuned models on
LiveMCPBench and transfers to unseen MCP servers.

\item \textbf{Identification of reasoning—tool seesaw and corresponding interference mitigation design.}
We characterize the reasoning tool seesaw as a joint-learning
failure mode and investigate disentangled token mask along with gradient surgery as
the potential mechanism.
Token-masked reward attribution separates direct supervision
for reasoning and action tokens, while bidirectional
gradient surgery mitigates residual interference through
shared parameters.
Experiments and ablations show that the full framework
achieves higher task success with shorter reasoning
than the untuned backbone and that each component contributes
(Section~\ref{sec:results}).

\item \textbf{Open resources for the community.}
We release 8{,}000 training samples curated from over
200{,}000 candidates through a synthesize-and-recheck
pipeline, together with training code and model weights,
to support reproducible research on MCP-native learning,
reasoning—tool interference, and generalizable agentic interaction.

\end{itemize}
\section{Related Work}
\label{sec:related}

Tool-augmented language models extend text generation through external computation and environment interaction. Toolformer~\citep{schick2023toolformer} learns API use by generating candidate tool calls and retaining those that improve subsequent token prediction. ReAct~\citep{yao2023react} interleaves reasoning, actions, and observations, enabling models to update their plans using environmental feedback. ToolLLM~\citep{qin2024toolllm} combines instruction tuning over 16{,}464 real-world APIs with tool retrieval and search-based solution generation, evaluating generalization to unseen tools, unseen categories, and an external API benchmark.

The Model Context Protocol~\citep{mcp,mcp2025tools} standardizes client--server interfaces for exposing capabilities and invoking tools. Server registries and community directories support the distribution and discovery of these capabilities. The official MCP Registry~\citep{mcp_registry} exposes standardized server metadata through a public API for downstream discovery services. Smithery~\citep{smithery} combines a searchable server registry with tooling for publishing servers and configuring client connections, while MCP.so~\citep{mcpso} organizes community-contributed servers through searchable listings, categories, and integration information. These collections make implementations spanning search, databases, developer tools, and other services easier to discover and reuse.
MCP-Zero~\citep{fei2025mcpzero} develops active tool discovery through capability requests, hierarchical semantic routing, and iterative toolset expansion. MCP-Flow~\citep{wang2026mcpflow} constructs large-scale instruction--call data and execution trajectories for supervised training and retrieval augmentation, with evaluations on unseen tools, unseen servers, and downstream agent tasks. Several benchmarks examine interaction with real MCP services. LiveMCPBench~\citep{mo2025livemcpbench} covers 70 servers and 527 tools across six domains and introduces a ReAct-based agent with routing, execution, and response actions in a partially observable environment. MCP-Universe~\citep{luo2025mcpuniverse} evaluates realistic tasks across multiple domains using execution-based evaluators, while MCP-Atlas~\citep{bandi2026mcp} emphasizes multi-step, cross-server workflows, tool selection among distractors, and claim-level assessment.

MCP also provides a common environment interface for agent-training infrastructure. RLFactory~\citep{chai2025rlfactory} decouples tool execution from policy training through MCP-based tool registration, asynchronous invocation, and flexible reward computation. OpenPipe's ART supports an MCP training pipeline that generates scenarios from tool descriptions, collects interaction trajectories, and applies RULER feedback for policy optimization~\citep{openpipe_mcprl}. MCP-Universe RL~\citep{luo2026mcpuniverserl} provides isolated environment provisioning, lifecycle management, and staged rollout orchestration, while integrating existing RL backends for policy updates. These systems illustrate how heterogeneous tools and task environments can be exposed through a shared interface and integrated into reusable training pipelines for multi-step workflows. ATLAS~\citep{gupta2026atlas} studies reinforcement fine-tuning for small language models operating in large MCP toolspaces. It combines incremental server and tool loading with programmatic tool orchestration to control context growth, and uses task-specific rubrics to provide structured training feedback.

\section{Methodology}
\label{sec:method}

\subsection{Thought-MDP Agent Logic}
\label{sec:agent_logic}
Figure~\ref{fig:pipe} illustrates the MCP-native interaction loop. Conventional tool calling selects a tool schema and generates its arguments. MCP additionally exposes a server-discovery layer: the agent routes a query to relevant server-wrapped capabilities, invokes their tools, and uses the returned observations to decide what to do next. The resulting policy must coordinate internal reasoning with external actions rather than generate a single function call from a fixed inventory.

\begin{figure*}[t]
    \centering
    \includegraphics[width=0.99\linewidth]{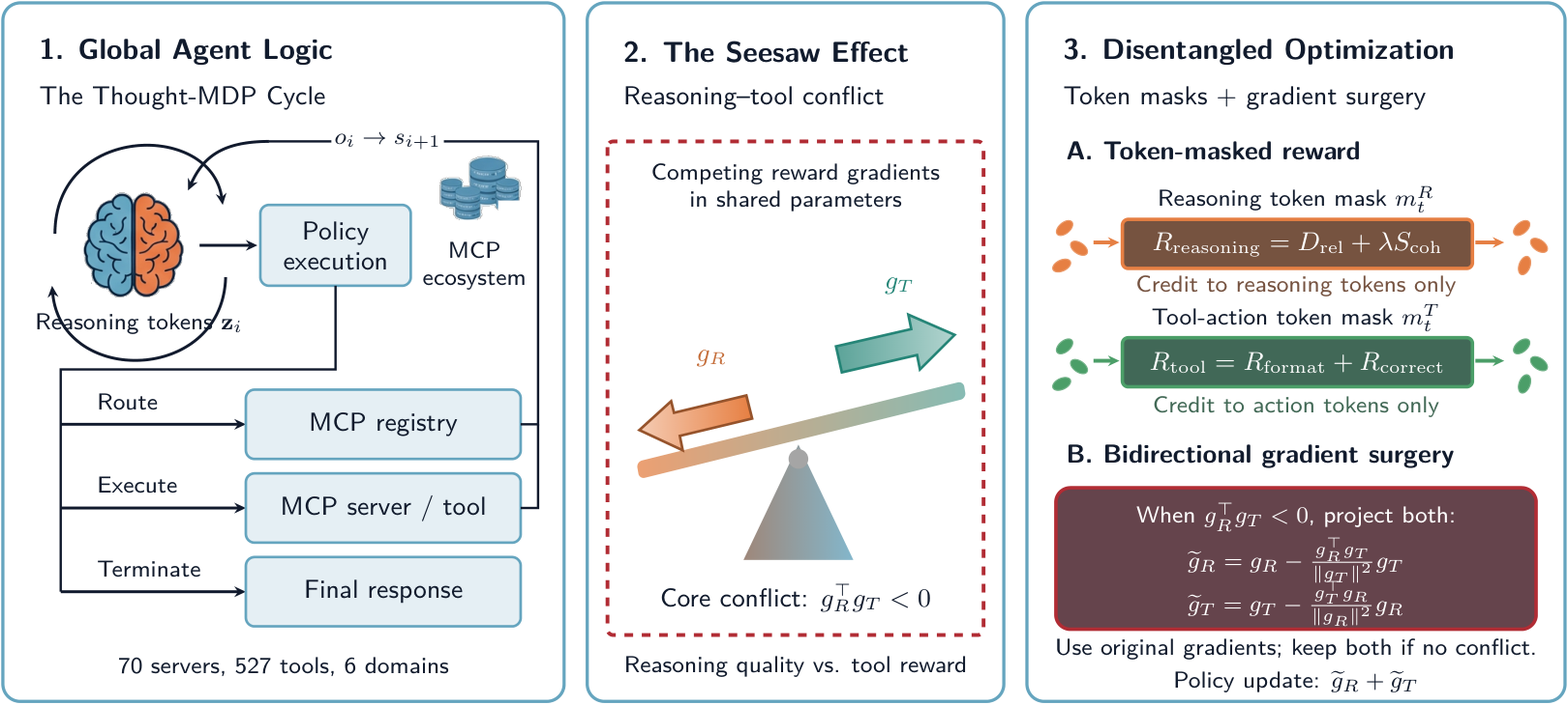}
    \caption{Overview of \modelname.}
    \label{fig:pipe}
\end{figure*}

We formalize this process as a \emph{Thought-MDP},
$\langle \mathcal{S},\mathcal{A},p,r,\gamma,\mathcal{T},\mathcal{C},p_{\mathcal{T}}\rangle$.
An environment state $s\in\mathcal{S}$ contains the user query, interaction history, and tool observations. The external action space $\mathcal{A}$ comprises three choices: \textsc{Route}, which retrieves relevant MCP servers or tools; \textsc{Execute}, which invokes a selected tool with JSON-formatted arguments; and \textsc{Terminate}, which returns a final response. The transition function $p:\mathcal{S}\times\mathcal{A}\rightarrow\Delta(\mathcal{S})$ incorporates action outcomes into the context, while $r:\mathcal{S}\times\mathcal{A}\rightarrow\mathbb{R}$ and $\gamma\in[0,1)$ specify external reward and discounting.

The additional thought state $t\in\mathcal{T}$ records explicit internal reasoning. Thought actions $c\in\mathcal{C}$ generate reasoning tokens without directly changing the environment, and $p_{\mathcal{T}}:\mathcal{S}\times\mathcal{T}\times\mathcal{C}\rightarrow\Delta(\mathcal{T})$ describes their evolution. A shared policy parameterized by $\theta$ generates both thought and external-action tokens. For a query $q$, its interaction trajectory is
\[
\tau=\bigl(\langle s_i,t_i,a_i,o_i\rangle\bigr)_{i=1}^{K},
\qquad K\leq L,
\]
where $L$ is the interaction limit, $o_i$ is the observation associated with action $a_i$, and $t_i=(z_{i,1},\ldots,z_{i,L_i})$ is the reasoning span preceding that action. For termination, no further tool observation is required. This distinction between thought and action spans provides the basis for separate reward attribution.

\subsection{Reward Components}
\label{sec:rewards}
We use two reward families: one scores the relevance and coherence of explicit reasoning, and the other scores the formatting and correctness of external actions. 

\minisection{L1: Reasoning quality reward}
The reasoning output is parsed for an explicit planning structure following the Harmony format~\citep{agarwal2025gpt}:
\begin{equation}
 r_1^{\text{preamble}}=
 \mathbb{I}\!\left[\text{STRUCT}(\mathbf{z})=
 \langle\text{preamble},\text{analysis},\text{action}\rangle\right],
\end{equation}
where $\mathbf{z}$ denotes the reasoning tokens for the query. The content reward combines relevance density and inter-sentence coherence.

\textbf{Relevance density.} KeyBERT extracts task-specific keywords from the input query to form a vocabulary $V_{\mathrm{rel}}$. For the reasoning text $T_{\mathrm{reasoning}}$, we compute
\begin{equation}
 D_{\mathrm{rel}}=
 \frac{\sum_{t\in T_{\mathrm{reasoning}}}\mathbb{I}(t\in V_{\mathrm{rel}})}
 {|T_{\mathrm{reasoning}}|}.
\end{equation}
This lexical proxy rewards task-related content relative to reasoning length rather than rewarding length itself.

\textbf{Coherence score.} We split the reasoning into $N$ sentences and append matched query keywords to each sentence before encoding it with a pretrained sentence embedding model~\citep{reimers2019sentence}. With sentence embeddings $e_1,\ldots,e_N$, the coherence score is
\begin{equation}
 S_{\mathrm{coh}}=
 \frac{1}{N-1}\sum_{i=1}^{N-1}
 \frac{\cos(e_i,e_{i+1})+1}{2}.
\end{equation}
The combined reasoning reward is
\begin{equation}
 R_{\mathrm{reasoning}}=D_{\mathrm{rel}}+\lambda S_{\mathrm{coh}},
 \label{eq:reasoning_reward}
\end{equation}
where $\lambda$ controls the coherence contribution. Keyword overlap and embedding continuity provide inexpensive shaping signals, but neither directly verifies logical correctness.

\minisection{L2: Tool syntax and correctness reward}
The action reward is $R_{\mathrm{tool}}=R_{\mathrm{format}}+R_{\mathrm{correct}}$. The format term rewards syntactically valid JSON, while the correctness term evaluates routing or execution against reference tool-call information.

\textbf{Execution.} For an execution action, the heuristic score is
\begin{equation}
\begin{aligned}
 H_{\mathrm{exec}}={}&R_{\mathrm{format}}+s_{\mathrm{base}}+s_{\mathrm{req}}
 +s_{\mathrm{srv}}+s_{\mathrm{tool}}\\
 &+\mathrm{IoU}_{\mathrm{keys}}+N_{\mathrm{val\_match}}.
\end{aligned}
\end{equation}
Here, $s_{\mathrm{base}}$ and $s_{\mathrm{req}}$ reward a parameter dictionary and the presence of required keys; $s_{\mathrm{srv}}$ and $s_{\mathrm{tool}}$ reward the correct server and tool; $\mathrm{IoU}_{\mathrm{keys}}$ measures overlap between predicted and reference argument keys; and $N_{\mathrm{val\_match}}$ counts exactly matched argument values. The correctness reward is rescaled according to the number of reference arguments:
\begin{equation}
 R_{\mathrm{correct}}^{\mathrm{exec}}
 =6\left(\frac{H_{\mathrm{exec}}}{2+|A_{\mathrm{gt}}|}\right)-3.
\end{equation}

\textbf{Routing.} A routing action is scored by the validity of its search query and whether its results recall the reference server and tool:
\begin{equation}
\begin{aligned}
 H_{\mathrm{route}}={}&R_{\mathrm{format}}+s_{\mathrm{base}}+s_{\mathrm{query}}\\
 &+s_{\mathrm{srv\_match}}+s_{\mathrm{tool\_match}},
\end{aligned}
\end{equation}
where the last two terms indicate successful recall. The corresponding correctness reward is
\begin{equation}
 R_{\mathrm{correct}}^{\mathrm{route}}
 =6\left(\frac{H_{\mathrm{route}}}{3}\right)-2.
\end{equation}
Severe formatting errors or missing required parameters incur penalties, such as $-2.0$ or $-3.0$. A well-formed final answer that successfully concludes the task receives a modest terminal shaping reward. Thus, although denoted $R_{\mathrm{tool}}$, this family also covers termination behavior. Reward hyperparameters are given in Appendix~\ref{app:setup}.

\subsection{Seesaw-Mitigated Tuning}
\label{sec:tuning}
Separate reward definitions do not by themselves prevent interference during joint optimization. We address this problem at two levels: token masking separates direct reward attribution, and bidirectional gradient surgery removes opposing components from the resulting shared-parameter updates.

\subsubsection{The Reasoning--Tool Seesaw}
\label{sec:seesaw}
Reasoning and tool use are complementary at inference time, but their training objectives need not improve together. We call an optimization trade-off in which progress on one objective is accompanied by deterioration on the other the \emph{reasoning--tool seesaw}. This is a failure mode of joint learning, not an inherent incompatibility between the two capabilities.

A straightforward baseline, \emph{integrated rewarding}, optimizes the scalar sum $R_{\mathrm{joint}}=R_{\mathrm{reasoning}}+R_{\mathrm{tool}}$. Aggregate improvement can then conceal deterioration in an individual component. Two forms of coupling are relevant. First, assigning the summed reward to all generated tokens gives reasoning and action tokens the same direct supervision despite their different roles. Second, even after separating these token-level signals, both streams update shared model parameters and can induce opposing directions.

We distinguish the behavioral trade-off from its possible optimization mechanism: a negative inner product between objective gradients indicates local interference, but does not alone establish that the two behavioral metrics move in opposite directions during training. 

\subsubsection{Token-Masked Disentangled Reward}
\label{sec:mask}
Let $y_{1:M}$ be the generated token sequence, $h_t$ the decoding history, and $A_t^R,A_t^T$ the advantages estimated from the reasoning and action reward families, respectively. We write the reward-driven policy-gradient terms below; the common RL optimization and reference-policy regularization settings are described in Appendix~\ref{app:setup}. Under integrated attribution,
\begin{equation}
 \nabla_\theta J_{\mathrm{int}}=
 \mathbb{E}\!\left[\sum_{t=1}^{M}
 \nabla_\theta\log\pi_\theta(y_t\mid h_t)
 (A_t^R+A_t^T)\right].
 \label{eq:integrated_grad}
\end{equation}

Instead, define disjoint masks $m_t^R,m_t^T\in\{0,1\}$ with $m_t^Rm_t^T=0$. The reasoning mask selects explicit thought spans. The action mask selects generated external-action tokens, including routing and execution calls, JSON arguments, and final-answer tokens receiving terminal shaping. User inputs and tool observations are not generated action tokens. The masked gradient is
\begin{equation}
\begin{aligned}
 \nabla_\theta J_{\mathrm{mask}}
 ={}&\underbrace{\mathbb{E}\!\left[\sum_{t=1}^{M}m_t^R
 \nabla_\theta\log\pi_\theta(y_t\mid h_t)A_t^R\right]}_{g_R} +\underbrace{\mathbb{E}\!\left[\sum_{t=1}^{M}m_t^T
 \nabla_\theta\log\pi_\theta(y_t\mid h_t)A_t^T\right]}_{g_T}.
\end{aligned}
\label{eq:mask_grad}
\end{equation}
This removes direct cross-stream reward attribution: tool rewards no longer directly weight reasoning-token likelihoods, and reasoning rewards no longer directly weight action-token likelihoods. It does not make the two capabilities independent, since action generation conditions on preceding reasoning and both streams share parameters.

\subsubsection{Gradient Surgery}
\label{sec:gradient_surgery}
Let $g_R$ and $g_T$ now denote minibatch estimates of the components in Equation~\ref{eq:mask_grad}, using the gradient-ascent convention. When $g_R^\top g_T<0$, an infinitesimal step along either component is a descent direction for the other local objective. Following the projection principle of PCGrad~\citep{yu2020gradient}, we remove the conflicting component from \emph{both} gradients:
\begin{equation}
\begin{aligned}
\tilde g_R&=
\begin{cases}
 g_R-\dfrac{g_R^\top g_T}{\|g_T\|_2^2}g_T,
 &g_R^\top g_T<0,\\[5pt]
 g_R,&\text{otherwise},
\end{cases}\\
\tilde g_T&=
\begin{cases}
 g_T-\dfrac{g_T^\top g_R}{\|g_R\|_2^2}g_R,
 &g_R^\top g_T<0,\\[5pt]
 g_T,&\text{otherwise},
\end{cases}\\
 g_{\mathrm{update}}&=\tilde g_R+\tilde g_T.
\end{aligned}
\label{eq:surgery}
\end{equation}
Both projections use the original gradients, not a sequentially modified pair. A zero gradient cannot satisfy the negative-inner-product condition and is left unchanged. The rule is symmetric: neither reasoning nor tool use is a privileged anchor.

On conflicting steps, $\tilde g_R^\top g_T=0$ and $\tilde g_T^\top g_R=0$ in exact arithmetic. These identities describe the local geometry of the reward-gradient intervention; they do not guarantee monotone improvement under finite optimizer steps, regularization, or stochastic estimation, nor do they require future unprojected gradients to remain compatible. When the original gradients are compatible, both are preserved.

Together, masking and surgery address distinct sources of interference without introducing an explicit length penalty. The reasoning reward encourages relevance and coherence, while the action reward encourages valid and correct execution. Whether this combination yields more successful tasks with less verbose reasoning is an empirical question evaluated in Section~\ref{sec:results}, rather than a consequence guaranteed by gradient projection.

\section{Experiments}

\subsection{Setup}
\textbf{Benchmark.} We evaluate our method on LiveMCPBench~\citep{mo2025livemcpbench}, a comprehensive evaluation framework designed to assess MCP-native agents in real-world scenarios, encompassing 70 MCP servers and 527 tools. The benchmark covers 95 diverse daily tasks across six domains: Office, Leisure, Travel, LifeStyle, Finance, and Shopping. For OOD test, we test the trained model on MCP-Atlas~\citep{bandi2026mcp} and GAIA~\citep{mialon2024gaia}.

\textbf{Training Data.} We synthesize and filter $8{,}000$+ samples out of $200{,}000$+ samples pool for training.


\textbf{Evaluation Metrics.} We adopt three metrics to measure both reasoning behavior and task-level performance: (1)~\textbf{Avg.\ CoT Tokens}: the average number of chain-of-thought tokens generated per query, reflecting reasoning verbosity; (2)~\textbf{Avg.\ Chain}: the average number of tool-calling steps per trajectory, reflecting planning depth; and (3)~\textbf{Success Rate}: the proportion of tasks successfully completed, as the primary performance indicator.

\textbf{Devices.} Training was done using a 8*A6000 device. Resources from Shanghai Innovation Center \footnote{https://www.sii.edu.cn/} were also used.

\subsection{Research Questions}
Our evaluation addresses three research questions:
\begin{itemize}[leftmargin=10pt]
    \item \textbf{(RQ1) Effectiveness.} Does ecosystem-scale training on real-server, long-horizon interactions under the Thought-MDP framework improve MCP-native task success over the base policy and existing tool-tuned models?
    \item \textbf{(RQ2) Optimization mechanisms.} How do token-masked reward attribution and bidirectional gradient surgery affect reasoning–tool interference, task success, and reasoning verbosity?
    \item \textbf{(RQ3) Generalization and composition.} To what extent does the learned policy transfer to unseen MCP servers and broader agentic tasks, and can it flexibly compose capabilities across servers in response to intermediate observations?
\end{itemize}

\subsection{Results}
\label{sec:results}

\begin{table*}[!h]
\centering
\caption{Comparison of task-level performance on LiveMCP. (Qwen3-4B)}
\label{tab:main_results}
\resizebox{0.99\linewidth}{!}{
\begin{tabular}{l|ccc|ccc|ccc|ccc}
\toprule
\multirow{2}{*}{\textbf{Domain}} 
    & \multicolumn{3}{c|}{\textbf{Base}} & \multicolumn{3}{c|}{\textbf{\modelname (-surgery)}} 
    & \multicolumn{3}{c|}{\textbf{\modelname (-mask)}} & \multicolumn{3}{c}{\textbf{\modelname}}\\
\cmidrule(lr){2-4}\cmidrule(lr){5-7}\cmidrule(lr){8-10}\cmidrule(lr){11-13}
    & \makecell{Avg.\ CoT\\ Tokens} & \makecell{Avg.\\ Chain} & \makecell{Success\\ Rate (\%)} 
    & \makecell{Avg.\ CoT\\ Tokens} & \makecell{Avg.\\ Chain} & \makecell{Success\\ Rate (\%)} 
    & \makecell{Avg.\ CoT\\ Tokens} & \makecell{Avg.\\ Chain} & \makecell{Success\\ Rate (\%)} 
    & \makecell{Avg.\ CoT\\ Tokens} & \makecell{Avg.\\ Chain} & \makecell{Success\\ Rate (\%)} \\
\midrule
Office                   & 2048.26                           & 10                           & 22.58         &2389.58 & 12.32&      25.81      & 2369.65                           & 11.26                        & 16.13                             & 1803.39                           & 14.52                        & 25.81                            \\
Leisure                  & 933.64                            & 5.07                         & 14.29       &1686.07 & 9.93&        35.71                        & 1585.14                           & 9.21                         & 35.71                             & 1349.21                           & 8.93                         & 21.43                            \\
Travel                   & 2815.92                           & 9.67                         & 8.33         & 2690.33& 13&  8.33                             & 2948.92                           & 11.17                        & 16.67                             & 2715.58                           & 17.17                        & 50                               \\
LifeStyle                & 1616.4                            & 8.6                          & 13.33         & 1812.73& 10.67&   26.67                           & 1868.2                            & 10.6                         & 40                                & 1331.17                           & 12.13                        & 13.33                            \\
Finance                  & 1733.37                           & 9.14                         & 7.14        & 2212.07& 13.79&  7.14                              & 2626.71                           & 13.29                        & 21.43                             & 1665.29                           & 16.29                        & 21.43                            \\
Shopping                 & 945.89                            & 5.67                         & 22.22        & 973.44& 7.22&    22.22                           & 985.44                            & 6.33                         & 22.22                             & 818.67                            & 8.11                         & 33.33                            \\ \midrule
Overall                  & 1761.97                           & 8.47                         & 15.79          & 2072.49& 11.53&     22.11                        & 2154.78                           & 10.67                        & 24.21                             & 1663.53                           & 13.31                        & 26.32                            \\ 
\bottomrule
\end{tabular}
}
\end{table*}


\begin{table}[!h]
    \centering
    \begin{minipage}{0.45\textwidth}
        \centering
        \caption{Performance under different modes.}
        \label{tab:left}
        \resizebox{0.99\linewidth}{!}{
        \begin{tabular}{cc|cc}
\hline
\multicolumn{2}{c|}{Method}               &  \begin{tabular}[c]{@{}c@{}}Avg.\\ Chain\end{tabular} & \begin{tabular}[c]{@{}c@{}}Success\\ Rate (\%)\end{tabular} \\ \hline
\multirow{2}{*}{Base}     & Reasoning Off   & 11.85                                                & 14.74                                                       \\
                          & Reasoning On & 8.47                                                 & 15.79                                                       \\ \hline
\multirow{2}{*}{MCPDisCo} & Reasoning Off  & 8.29                                                 & 16.84                                                       \\
                          & Reasoning On & 13.31                                                & 26.32                                                       \\ \hline
\end{tabular}
}
    \end{minipage}
    \hfill
    \begin{minipage}{0.4\textwidth}
        \centering
        \caption{Performance on LiveMCP}
        \label{tab:ood}
    \resizebox{0.99\linewidth}{!}{
    \begin{tabular}{c|ccc}
    \hline
    Model                & \makecell{Avg.\ CoT\\ Tokens} & \makecell{Avg.\\ Chain} & \makecell{Success\\ Rate (\%)}  \\ \hline
    Base                 & 1761.97     & 8.47 & 15.79\\
    ToolACE-8B           & 0&  9.08 & 8.42     \\
    Hammer-7B            & 255.67 &   12.05 & 12.63    \\
    \modelname     & 1663.53 &  13.31 & 26.32    \\ \hline
    \end{tabular}
    }
    \end{minipage}
\end{table}

\begin{figure}[!h]
    \centering
    \begin{subfigure}[b]{0.48\textwidth}
        \centering
        \includegraphics[width=1.0\linewidth]{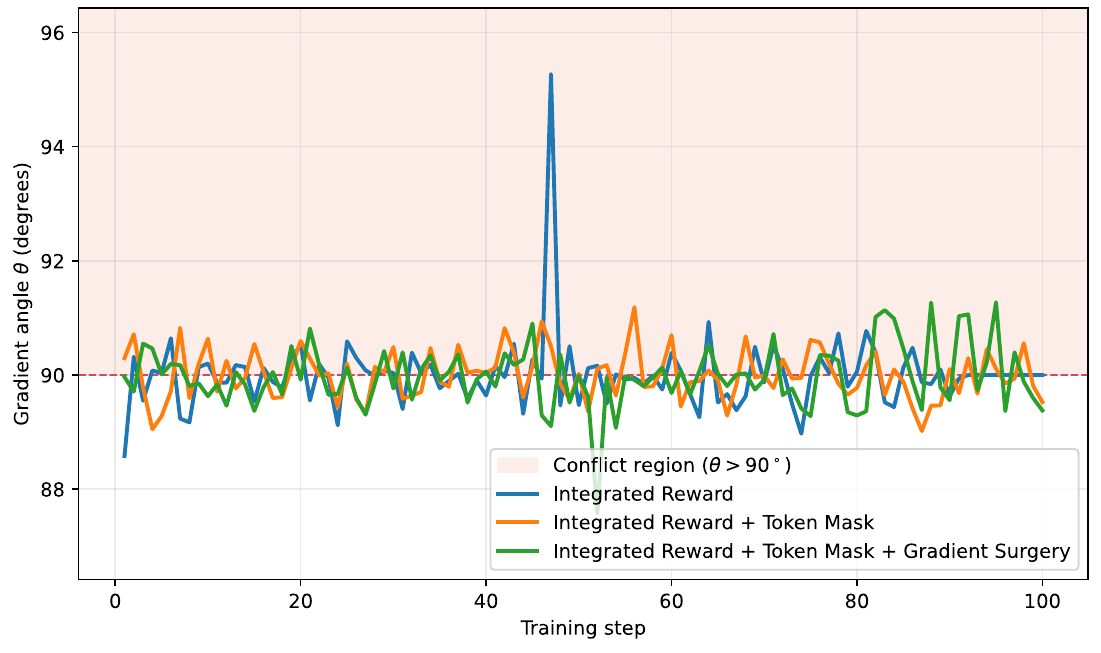}
        \caption{Gradient-angle trajectories between the reasoning and tool policy gradients.}
        \label{fig:grad_trajectory}
    \end{subfigure}
    \hfill 
    \begin{subfigure}[b]{0.48\textwidth}
        \centering
        \includegraphics[width=1.0\linewidth]{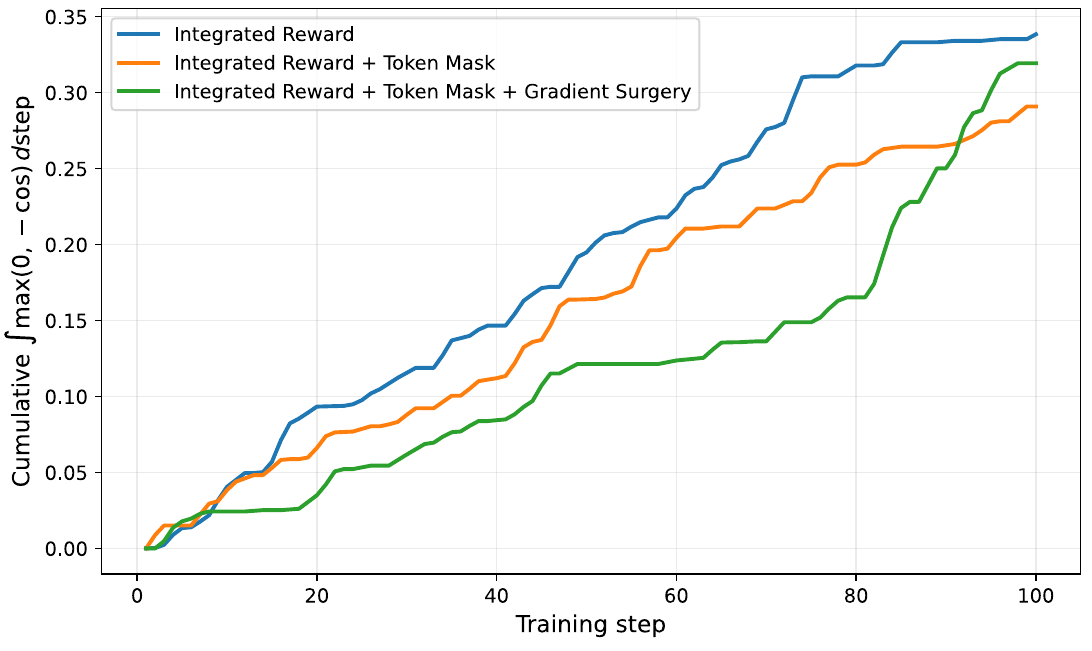}
        \caption{The cumulative negative-cosine integral.}
        \label{fig:cosine}
    \end{subfigure}
    \caption{Illustrations of the seesaw effect.}
    \label{fig:two_images}
\end{figure}
    \paragraph{Main Result (RQ1)}Table 1 reports the task-level performance of the base agent the agents trained with different rewarding. Overall, reinforcement learning with \modelname improves the success rate from 15.79\% to 26.32\%, corresponding to an absolute gain of 10.53 percentage points compared to the base model without tuning. This result indicates that MCP-native tool use benefits substantially from the explicit optimization and disentanglement of routing, execution, and termination behaviors.
    \paragraph{Seesaw-Mitigation (RQ2)} 
    In Figure~\ref{fig:grad_trajectory}, all three trajectories fluctuate near the orthogonality boundary. Integrated rewarding exhibits the largest isolated excursion into the conflict region, while the masked configurations show smaller excursions. The pattern is consistent with intermittent shared-parameter interference, rather than a sustained interval in which one behavioral capability necessarily deteriorates as the other improves.

Figure~\ref{fig:cosine} provides a complementary view. Both masked configurations end the displayed window below integrated rewarding in cumulative angular conflict. The full method accumulates less conflict than masking alone through much of the early and middle portion of the window, but rises above masking alone near the end. Thus, the displayed curves do not support a claim that adding surgery monotonically lowers accumulated conflict throughout training or yields the lowest final value. Its pointwise projection rule and the evolution of gradients encountered along a training trajectory are distinct: if the curves record gradients before projection, future negative alignment remains compatible with applying surgery at each update. The measurement stage must be established before interpreting these curves as a direct check of the projection itself.

    \paragraph{Reasoning Density and Token Efficiency (RQ2)}

A central prediction of the seesaw analysis in Section~\ref{sec:seesaw} is that disentangled rewarding should not merely trade quality for quantity in the reasoning stream, but should raise reasoning \emph{density}---the amount of task-relevant content per token—thereby enabling shorter chain-of-thought at equal or higher success. The results in Table~\ref{tab:main_results} confirm this prediction along three orthogonal axes:

\begin{itemize}[leftmargin=10pt]
\item \textbf{Shorter CoT than the untuned baseline.} Disentangled rewarding reduces average CoT tokens to $1663.53$, \emph{below} the $1761.97$ produced by the untuned base agent. This is unusual: RL fine-tuning for reasoning typically inflates CoT length (as does our own integrated-rewarding baseline, which reaches $2154.78$). \modelname is the only configuration that improves success while shrinking the reasoning budget.
\item \textbf{Deeper tool interaction.} Simultaneously, the average tool-call chain grows from $8.47$ to $13.31$. Tokens are being reallocated from redundant reasoning into productive tool invocations—consistent with the mechanism proposed in Section~\ref{sec:seesaw}: separating credit for $R_{\text{reasoning}}$ and $R_{\text{tool}}$ lets the policy stop over-verbalizing in service of a shared scalar.
\item \textbf{Efficiency $\times$ effectiveness.} Combining the two, \modelname delivers a $10.53$-point success-rate lift ($15.79\% \to 26.32\%$) using $5.6\%$ fewer CoT tokens than the base model, giving a strictly better operating point on the (tokens, success) frontier than either the base or the integrated-rewarding variant.
\end{itemize}

\begin{figure}[!h]
    \centering
    \includegraphics[width=0.5\linewidth]{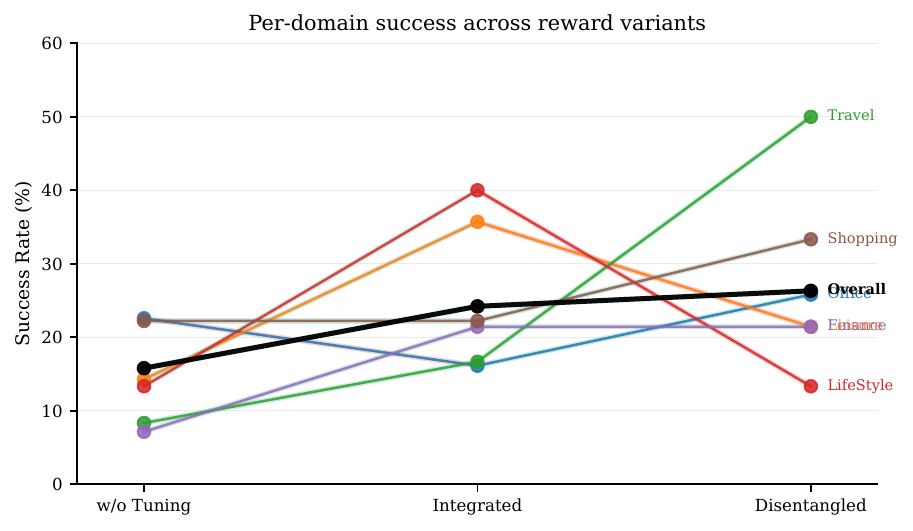}
    \caption{Per-domain change in average CoT tokens vs.\ change in success rate.}
    \label{fig:domain}
\end{figure}

Figure~\ref{fig:domain} decomposes this behavior across the six LiveMCPBench domains. The clear majority of domains lie in the ``fewer tokens, higher success'' quadrant under disentanglement, corroborating that the aggregate efficiency gain is not driven by a single outlier domain.


\paragraph{Out-of-Distribution Generalization (RQ3)}
\label{sec:ood}

To probe whether the learned reasoning—tool interaction policy—transfers beyond the training server distribution, we evaluate on MCP-Atlas~\citep{bandi2026mcp}, a benchmark whose server population is disjoint from LiveMCPBench, and a general agent benchmark GAIA~\citep{mialon2024gaia}. Results are summarized in Table~\ref{tab:atlas} and Table~\ref{tab:gaia}, from which we can see that the learned policy generalizes to unseen servers and broader agentic tasks.

\begin{table}[!h]
    \centering
    \begin{minipage}{0.45\textwidth}
        \centering
        \caption{OOD test on MCP-Atlas}
    \label{tab:atlas}
    \begin{tabular}{c|cc}
    \hline
    Model                & ACC    & \makecell{Mean Coverage} \\ \hline
    Base                 & 0.0186 & 0.1165        \\
    ToolACE-8B           & 0.0268 & 0.1258      \\
    Hammer-7B            & 0.0205 & 0.1194      \\
    \modelname     & 0.0404 & 0.1360        \\ \hline
    \end{tabular}
    \end{minipage}
    \hfill
    \begin{minipage}{0.45\textwidth}
        \centering
        \caption{OOD test on GAIA}
        \label{tab:gaia}
        \begin{tabular}{c|cc}
        \hline
        Model                & L1   & L2 \\ \hline
        Base                 & 0.1320 &    0.1050  \\
        ToolACE-8B           & 0.0430 &  0.0230     \\
        Hammer-7B            & 0.0750 &  0.0470     \\
        \modelname     & 0.1510 &  0.0700    \\ \hline
        \end{tabular}
    \end{minipage}
\end{table}

Moreover, \modelname shows superior performance on the competing benchmarks over tool-calling models ToolACE-8b~\cite{liu2025toolace} and Hammer-7B~\cite{lin2024hammer}, further demonstrating the capability of \modelname. We illustrate this generalization hierachy as below:

\begin{figure}[!h]
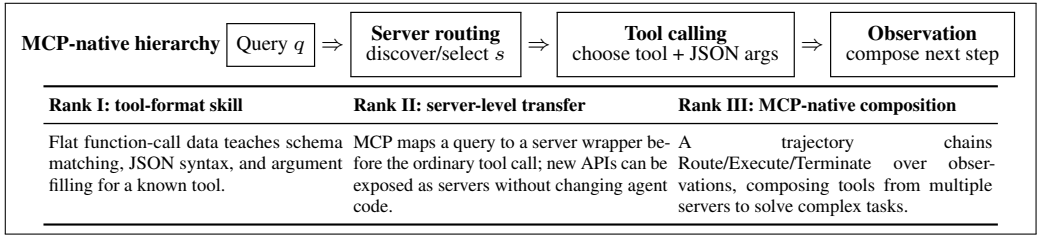

    \centering
    \scriptsize
    \setlength{\fboxsep}{3pt}
    \setlength{\tabcolsep}{2pt}
    \renewcommand{\arraystretch}{1.08}
    \fbox{%
    \begin{minipage}{0.96\linewidth}
    \centering
    \resizebox{0.98\linewidth}{!}{%
    \begin{tabular}{@{}c@{\;}c@{\;}c@{\;}c@{\;}c@{\;}c@{\;}c@{\;}c@{}}
    \textbf{MCP-native hierarchy} &
    \fbox{\strut Query $q$} &
    $\Rightarrow$ &
    \fbox{\begin{tabular}{c}\textbf{Server routing}\\[-2pt] discover/select $s$\end{tabular}} &
    $\Rightarrow$ &
    \fbox{\begin{tabular}{c}\textbf{Tool calling}\\[-2pt] choose tool + JSON args\end{tabular}} &
    $\Rightarrow$ &
    \fbox{\begin{tabular}{c}\textbf{Observation}\\[-2pt] compose next step\end{tabular}}
    \end{tabular}}
    \vspace{2pt}

    \begin{tabular}{p{0.29\linewidth}p{0.31\linewidth}p{0.31\linewidth}}
    \toprule
    \textbf{Rank I: tool-format skill} &
    \textbf{Rank II: server-level transfer} &
    \textbf{Rank III: MCP-native composition}\\
    \midrule
    Flat function-call data teaches schema matching, JSON syntax, and argument filling for a known tool. &
    MCP maps a query to a server wrapper before the ordinary tool call; new APIs can be exposed as servers without changing agent code. &
    A trajectory chains Route/Execute/Terminate over observations, composing tools from multiple servers to solve complex tasks.\\
    \bottomrule
    \end{tabular}
    \end{minipage}}
    \vspace{-4pt}
    \caption{Hierarchy behind MCP-native generalization. Ordinary tool-call tuning covers the lower tool-format layer; MCP adds a server-routing and wrapping layer, enabling flexible, composable functions and a higher generalization rank than flat function calling.}
    \label{fig:pipe}
    \vspace{-6pt}
\end{figure}

Overall, our real-world, hierarchical (query -> server -> tool), long-horizon (>10 interaction turns) agent data, together with our disentangled and reasoning-focused tuning algorithm, makes the resulting tuned model generalizable for wide agent tasks. OOD tests on common and untuned benchmarks validate the conclusion.

We therefore position MCP as a shared interaction substrate for the agentic web, where capabilities exposed by heterogeneous servers can be discovered, invoked, and composed into extended workflows. Realizing this potential requires policies that generalize beyond individual tools and fixed task compositions. To this end, \modelname learns from real-server, long-horizon MCP interaction data organized around a hierarchy of tool-format skills, server-level capabilities, and cross-server compositions. This structure exposes reusable capabilities at multiple
levels, while flexible task composition places them in diverse workflows requiring sustained, observation-conditioned coordination.


\section{Conclusion}
We presented \modelname, an ecosystem-scale RL framework that positions MCP as a shared interaction substrate for the agentic web. Under a Thought-MDP formulation, \modelname learns to coordinate reasoning, server discovery, and tool execution through real-server, long-horizon interactions. 
The resulting training data spans a capability hierarchy from tool-format skills to server-level transfer and cross-server composition, enabling flexible construction of complex tasks beyond fixed tool inventories. Improved performance on unseen servers in MCP-Atlas and transfer to broader agentic tasks in GAIA support the generalization potential of this approach. 
These findings suggest that MCP provides not only interface standardization but also a compositional foundation for learning reusable interaction policies.

To address the reasoning--tool seesaw, \modelname combines token-masked reward attribution with bidirectional gradient surgery, mitigating interference at both the token and shared-parameter levels. On LiveMCPBench, the full method improves success from 15.79\% to 26.32\% while reducing average chain-of-thought tokens by 5.6\% relative to the untuned base. Ablations support the complementary benefits of the two mechanisms, and gradient diagnostics provide evidence of optimization interference without implying its complete elimination. Together, these results highlight the value of combining hierarchical MCP interaction data with interference-aware optimization to develop agents that discover, invoke, and recombine capabilities across the agentic web. We release training data, code, and model weights to support further research.

\subsubsection*{Reproducibility statement}
Complete optimization, rollout, infrastructure, and evaluation settings are provided in Appendix~\ref{app:setup}. We also provide anonymous code and model artifacts in the supplementary material. Because the reported live-server runs are expensive and server behavior can change over time, we explicitly identify single-run reporting and live-infrastructure variability as limitations.

\subsubsection*{Ethics statement}
Training and evaluating agents against live MCP servers creates privacy, security, and unintended-side-effect risks. Our experiments use benchmark tasks and vetted servers, avoid private user data, cap interaction length, and treat tool outputs as untrusted observations. Released trajectories should be screened for credentials and personally identifiable information, and deployment should use least-privilege credentials, sandboxing, human confirmation for consequential actions, and auditable execution logs.

\subsection*{AI use statement}
Generative AI tools were used to assist with language editing, literature discovery, and LaTeX-format migration. All technical claims, citations, equations, experimental values, and generated edits were reviewed by the authors, who take responsibility for the final content. No AI-generated experimental result is presented as an observed result.

\bibliography{ref_revised}
\bibliographystyle{iclr2027_conference}
\appendix
\section{Implementation Details}

\label{app:setup}

\minisection{Backbone and RL algorithm}
All \modelname variants and the untuned baseline start from the same publicly released checkpoint, \texttt{Qwen3-4B}. The RL fine-tuning objective is optimized with a GRPO~\citep{shao2024deepseekmath}-style policy-gradient update in which the KL to a frozen reference policy is added to the loss as a soft constraint. The advantage estimator is a group-relative baseline: for each prompt we roll out $G=8$ trajectories in parallel and take the group-mean reward as the value baseline, which sidesteps the need for a separate critic head.

\minisection{Optimization}
We use AdamW with $\beta_1=0.9,\beta_2=0.95$, weight decay $0.01$, a peak learning rate of $1\times 10^{-6}$, and a cosine schedule with $5\%$ linear warmup. The global batch size is $32$ prompts (i.e., $32\times G = 256$ trajectories per update) and we train for $512$ policy-update steps, which corresponds to roughly one and a half epochs over the $8{,}000$-sample training set. The KL coefficient to the reference policy is $\beta_{\mathrm{KL}}=1\times 10^{-3}$. All variants use the same optimizer and schedule.

\minisection{Reward hyperparameters}
The reasoning reward weight is $\lambda=0.5$ (Eq.~4 of the main paper), balancing $D_{\mathrm{rel}}$ and $S_{\mathrm{coh}}$. Sentence embeddings for $S_{\mathrm{coh}}$ are computed with the \texttt{all-MiniLM-L6-v2} encoder~\citep{reimers2019sentence}; task keywords for $D_{\mathrm{rel}}$ are extracted with KeyBERT (top-$10$ n-grams, $n\!\in\!\{1,2\}$). For the tool syntax reward, the format base is $R_{\mathrm{format}}\!=\!1$, and the additive bonuses $s_{\mathrm{base}},s_{\mathrm{req}},s_{\mathrm{srv}},s_{\mathrm{tool}}$ are each set to $1$; severe formatting errors receive a $-2.0$ penalty and missing required keys receive a $-3.0$ penalty (main paper, \S3.3).

\minisection{Rollout and decoding}
Rollouts use sampling with temperature $0.9$ and $\mathrm{top}\text{-}p\!=\!0.95$; the per-trajectory token budget is $8192$ tokens, of which up to $4096$ may be consumed by chain-of-thought. Each trajectory has a hard cap of $20$ tool-invocation steps to bound MCP-server exposure per query. Tool servers are pinned to the same 70-server pool used at evaluation to ensure a consistent action distribution.

\minisection{Infrastructure and runtime}
Training is performed on a single node with $8\times$ NVIDIA A600-80GB GPUs. Rollouts are sharded across the GPUs and dispatch to MCP servers over a local network. A full $512$-step run takes approximately $36$ hours end-to-end, with roughly $70\%$ of wall-clock spent on MCP tool invocations rather than gradient updates.

\minisection{Evaluation protocol}
All numbers in Tables~\ref{tab:main_results}~and~\ref{tab:ood} are computed at greedy decoding (temperature $0$) on the official LiveMCPBench task splits and the MCP-Atlas, GAIA OOD split, with a $20$-step tool-invocation cap and a $8192$-token generation budget matching the training setup. The Success Rate is defined by the benchmark's official judge; Avg.\ CoT Tokens and Avg.\ Chain are computed only over completed trajectories.

\end{document}